\documentclass[10pt,conference]{IEEEtran}
\usepackage{booktabs}						% professional-quality tables
\usepackage{multirow}						% tabular cells spanning multiple rows
\usepackage{amsfonts}						% blackboard math symbols
\usepackage{graphicx}						% figures
\usepackage{duckuments}		
\usepackage[caption=false,font=footnotesize,labelfont=sf,textfont=sf]{subfig}
\usepackage{amsfonts,amsmath}
\usepackage{xcolor,colortbl} 
\usepackage{makecell}
\usepackage{soul}
\usepackage{url}

\ifCLASSINFOpdf
\else
\fi
\begin{document}
%
% paper title
% Titles are generally capitalized except for words such as a, an, and, as,
% at, but, by, for, in, nor, of, on, or, the, to and up, which are usually
% not capitalized unless they are the first or last word of the title.
% Linebreaks \\ can be used within to get better formatting as desired.
% Do not put math or special symbols in the title.
\title{CAARL: In-Context Learning for Interpretable Co-Evolving Time Series Forecasting}
%
%
% author names and IEEE memberships
% note positions of commas and nonbreaking spaces ( ~ ) LaTeX will not break
% a structure at a ~ so this keeps an author's name from being broken across
% two lines.
% use \thanks{} to gain access to the first footnote area
% a separate \thanks must be used for each paragraph as LaTeX2e's \thanks
% was not built to handle multiple paragraphs
%

\author{
Etienne~Tajeuna,~\IEEEmembership{Member,~IEEE,}
Patrick~Asante~Owusu,~\IEEEmembership{Fellow,~OSA,}
Armelle~Brun,~\IEEEmembership{Fellow,~OSA,}
and~Shengrui~Wang,~\IEEEmembership{Life~Fellow,~IEEE}% <-this % stops a space
\thanks{M. Shell was with the Department
of Electrical and Computer Engineering, Georgia Institute of Technology, Atlanta,
GA, 30332 USA e-mail: (see http://www.michaelshell.org/contact.html).}% <-this % stops a space
\thanks{J. Doe and J. Doe are with Anonymous University.}% <-this % stops a space
\thanks{Manuscript received April 19, 2005; revised August 26, 2015.}\\
\{etienne.gael.tajeuna, patrick.asante.owusu, shengrui.wang\}@usherbrooke.ca\\

}

% note the % following the last \IEEEmembership and also \thanks - 
% these prevent an unwanted space from occurring between the last author name
% and the end of the author line. i.e., if you had this:
% 
% \author{....lastname \thanks{...} \thanks{...} }
%                     ^------------^------------^----Do not want these spaces!
%
% a space would be appended to the last name and could cause every name on that
% line to be shifted left slightly. This is one of those "LaTeX things". For
% instance, "\textbf{A} \textbf{B}" will typeset as "A B" not "AB". To get
% "AB" then you have to do: "\textbf{A}\textbf{B}"
% \thanks is no different in this regard, so shield the last } of each \thanks
% that ends a line with a % and do not let a space in before the next \thanks.
% Spaces after \IEEEmembership other than the last one are OK (and needed) as
% you are supposed to have spaces between the names. For what it is worth,
% this is a minor point as most people would not even notice if the said evil
% space somehow managed to creep in.

% The paper headers
\markboth{Journal of \LaTeX\ Class Files,~Vol.~14, No.~8, August~2015}%
{Shell \MakeLowercase{\textit{et al.}}: Bare Demo of IEEEtran.cls for IEEE Journals}
% The only time the second header will appear is for the odd numbered pages
% after the title page when using the twoside option.
% 
% *** Note that you probably will NOT want to include the author's ***
% *** name in the headers of peer review papers.                   ***
% You can use \ifCLASSOPTIONpeerreview for conditional compilation here if
% you desire.

% If you want to put a publisher's ID mark on the page you can do it like
% this:
%\IEEEpubid{0000--0000/00\$00.00~\copyright~2015 IEEE}
% Remember, if you use this you must call \IEEEpubidadjcol in the second
% column for its text to clear the IEEEpubid mark.

% use for special paper notices
%\IEEEspecialpapernotice{(Invited Paper)}

% make the title area
\maketitle

% As a general rule, do not put math, special symbols or citations
% in the abstract or keywords.
\begin{abstract}
In this paper, we investigate forecasting co-evolving time series that feature intricate dependencies and non-stationary dynamics by using an LLM (Large Language Models) approach. We propose a novel modeling approach, named {\it Context-Aware AR-LLM} (CAARL), that provides an interpretable framework to decode the contextual dynamics influencing changes in co-evolving series. CAARL decomposes time series into autoregressive segments, constructs a temporal dependency graph, and serializes this graph into a narrative to allow processing by LLM. This design yields a chain-of-thought–like reasoning path, where intermediate steps capture contextual dynamics and guide forecasts in a transparent manner. By linking prediction to explicit reasoning traces, CAARL enhances interpretability while maintaining accuracy. Experiments on real-world datasets validate its effectiveness, positioning CAARL as a competitive and interpretable alternative to state-of-the-art forecasting methods.
\end{abstract}

% Note that keywords are not normally used for peerreview papers.

% For peer review papers, you can put extra information on the cover
% page as needed:
% \ifCLASSOPTIONpeerreview
% \begin{center} \bfseries EDICS Category: 3-BBND \end{center}
% \fi
%
% For peerreview papers, this IEEEtran command inserts a page break and
% creates the second title. It will be ignored for other modes.
\IEEEpeerreviewmaketitle

\section{Introduction}

The autoregressive (AR) model is a fundamental approach for univariate time series forecasting. In an $AR(p)$ process, each observation $s^i_t$ of a time series $S^i$ is expressed as a linear combination of its $p$ past values:  
\begin{align}\label{Eq:autoregression}
  s^i_t &= f(t|\Theta^i,(s^i_{t-l}, 1\leq l \leq p)) 
  = \sum_{l=1}^p \omega^i_l s^i_{t-l} + \epsilon_t ,
\end{align}
\noindent where $\Theta^i = \{\omega^i_l,\, 1\leq l \leq p,\, p\in \mathbb{N}^*\}$ are regression coefficients, and $\epsilon_t \sim \mathcal{N}(0,\,\sigma^2)$ is a Gaussian noise term. Owing to its interpretability, $AR(p)$ has been widely used for stationary processes. However, real-world time series are often non-stationary and evolve dependently or interact on each other,
%influenced by interactions among multiple series,
limiting the effectiveness of classical AR models.  

Such interactions arise in \textit{co-evolving time series}, where multiple synchronous sequences $\mathcal{CS} = \{S^i,\, i=1,2,\ldots,n\}$ capture the dynamics of some complex systems \cite{bai2020entropic, chen2018neucast}. This paradigm has gained attention in diverse domains, including traffic forecasting \cite{guo2019attention, huang2020lsgcn}, financial prediction \cite{salinas2019high, wang2021coupling}, and healthcare monitoring \cite{vosoughi2020large, das2021schizophrenia}. Modeling these processes requires capturing both individual series dynamics and their cross-dependencies.  

Non-linear and time-varying models have been proposed to address non-stationarity \cite{sen2019think, le2019shape}, but their complexity often undermines interpretability. Transparent models, such as linear regression or decision trees, facilitate understanding \cite{linardatos2020explainable} but may fail when co-evolving dependencies induce heterogeneous non-stationarity across series. This underscores the necessity of approaches that explicitly model temporal interactions and contextual dependencies.  

In this regard, \textit{in-context learning} provides a promising direction for capturing correlations and regime shifts in co-evolving time series \cite{jin2023time, chang2023llm4ts}. By leveraging contextual patterns rather than relying solely on historical autoregression, it enhances adaptability in dynamic environments such as finance, healthcare, and industrial systems. For instance, stock price forecasting benefits not only from sequential history but also from exogenous factors like news events, demanding context-aware models capable of handling evolving dependencies. 

{\bf The code will be made available upon acceptance to support reproducibility.}

\section{Overview} \label{motivation}
Methods have been developed to capture in-context information, including dependencies and interactions among time series while considering their sequential nature \cite{wu2020connecting, jin2022multivariate}. These approaches aim to clarify how different series influence each other, particularly when multiple time series evolve concurrently. Despite advancements in understanding the complex dynamics of co-evolving time series, many methods remain ``black boxes", offering accurate time series forecasting without explanations. This lack of clarity limits practical utility and trust. Moreover, due to their fixed interpretation of data relationships, these models struggle to adapt to new trends and shifts in non-stationary environments.

Given these challenges in forecasting co-evolving time series, a crucial scientific question arises: \textbf{Can we develop a model that is simple, straightforward, transparent, and capable of effectively handling the complexities of co-evolving time series data while providing reliable and interpretable explanations for its predictions?} This inquiry is fundamental to improving trust and understanding in practical applications where multiple time series co-evolve.

\begin{figure*}[!t]
    \centering
    \includegraphics[width=.75\textwidth]{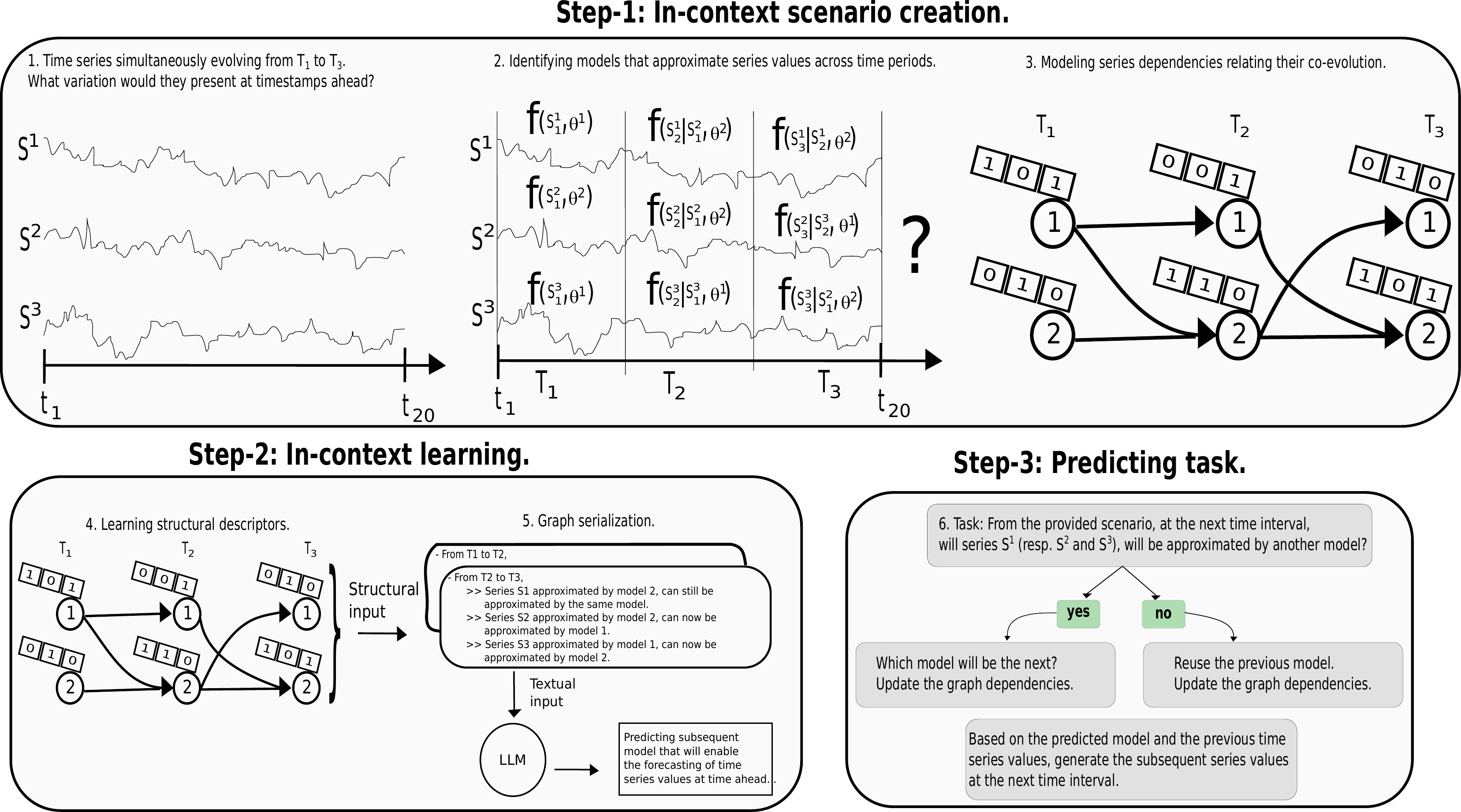}
    \caption{Overview of the proposed approach.}
    \label{fig:proposed}
\end{figure*}

\subsection{Overview of CAARL}

To tackle the aforementioned question, we propose {\it Context-Aware AR-LLM} (CAARL), a three-step framework for forecasting co-evolving time series (Fig.~\ref{fig:proposed}). Consider three series $S^1$, $S^2$, and $S^3$ observed from $t_1$ to $t_{20}$, with the goal of predicting beyond $t_{20}$. Each segment can be approximated by a simple autoregressive model, primarily characterized by two parameter sets, $\Theta^1$ and $\Theta^2$, across intervals $T_1$, $T_2$, and $T_3$. The shared use of the same model across different series and intervals—for example $S^1$ at $T_1$ and $S^3$ at $T_2$ 
% (\textcolor{red}{This is not true according to your figure (at Step1). According to your figure, $S^3$ at $T_1$ uses $\Theta^1$ while $S^1$ at $T_2$ uses $\Theta^2$. The coding at the third sub-figure (temporal graph) of Step-1 is also wrong at the $T_1$)} 
both using $\Theta^1$—reveals latent dependencies.

These dependencies are summarized in a temporal graph, where nodes represent autoregressive models and their associated series. The graph forms the \textit{in-context scenario}, a structural representation of evolving interrelations. In Step~2, the graph is serialized into a sequence describing how models are reused across intervals. In Step~3, an LLM reasons over this serialized representation to select the most suitable model for forecasting future values.

In short, CAARL combines interpretable autoregression with LLM-based reasoning: autoregressive models capture local dynamics, the graph encodes dependencies, and the LLM leverages these structures to produce accurate, interpretable forecasts.

\subsection{Why LLM?} The motivation of using an LLM in CAARL for co-evolving time series forecasting lies in its ability to enhance both interpretability and adaptability. Traditional methods, though effective, often function as ``black boxes" that provide accurate predictions without explaining the underlying relationships, limiting trust and practical utility. By integrating an LLM, CAARL transforms complex time series dependencies into human-readable narratives, offering clear explanations of how the series interact over time. Additionally, the LLM's dynamic learning capability ensures adaptability in non-stationary environments, allowing the model to adjust to changing patterns and trends. This makes CAARL both transparent and flexible, providing accurate forecasts while addressing the limitations of existing methods.

% The use of an LLM is crucial because it provides explanations in a human-readable format, enhancing user comprehension of the predictions. Additionally, the LLM's dynamic learning ability ensures it stays relevant and accurate, even in rapidly changing scenarios, by providing explanations that align with the latest data influences.

\subsection{Contributions} \label{contribution}
The major contributions of this paper are as follows:

$\bullet$ We present an interpretable approach designed to clarify the contextual dynamics driving changes in synchronized time series. Leveraging the clarity of straightforward autoregression models combined with a Large Language Model (LLM), this approach not only predicts future variations in the series but also offers insights into the underlying factors driving these changes.

$\bullet$ We construct a temporal graph to depict the dependencies that time series may exhibit at different timestamps. This graph encapsulates contextual information, showcasing the likely evolution of the series over time. Unlike conventional approaches that build graphs primarily based on sequence similarities, the temporal graph proposed emphasizes the essential pattern dependencies that drive the repetitive behavior inherent in co-evolving time sequences.

$\bullet$ We introduce a novel graph serialization principle that transforms the temporal graph into natural language text. This innovative technique enables us to articulate the graph's complex relationships coherently and understandably. By generating textual descriptions from the causal graph, we unveil comprehensive insights into time series' alterations over time.

$\bullet$ We validate the whole approach by implementing it on real-world data sets. We illustrate the strength of the proposed method in predicting the subsequent patterns the co-evolving time series may present at unseen times ahead.

\section{Related Works}
Recent advances in ensemble learning, graph-based modeling, causal inference, and Natural Language Processing (NLP) have substantially advanced time series forecasting. However, existing approaches still face limitations in jointly modeling co-evolving temporal dependencies, preserving interpretability, and enabling adaptive learning in dynamic and in-context settings—challenges directly addressed by our work.

\textbf{Ensemble models.} Ensemble methods improve forecasting robustness by aggregating diverse predictors, including random subspace approaches~\cite{stefenon2022time}, bootstrap aggregation~\cite{lee2020bootstrap}, and boosting techniques~\cite{sauer2022extreme}. While effective in reducing variance, these methods offer limited interpretability and treat temporal interactions implicitly, making them ill-suited for modeling explicit co-evolving dependencies across time series~\cite{ge5104377rethinking}. In contrast, our approach explicitly models temporal interdependencies through structured representations rather than opaque model aggregation.

\textbf{Graph-based models.} Graph-based forecasting frameworks leverage Graph Neural Networks (GNNs) to represent interactions among multivariate time series~\cite{cini2023graphdeeplearningtime, velivckovic2023everything, bacciu2020gentle}, with extensions capturing causal relations~\cite{xu2019scalable}. Recent advances move toward unified spatio-temporal modeling: FourierGNN introduces hypervariate graphs in the frequency domain for expressive learning~\cite{yi2024fouriergnn}, whereas TPGNN emphasizes interpretability via temporal polynomial graphs but depends on predefined adjacency structures~\cite{liu2022multivariate}. These design choices either sacrifice interpretability or constrain adaptability. Our work departs from fixed or frequency-driven graph constructions by learning dynamic dependency structures that evolve with the observed context.

\textbf{Causal-based models.} Causal discovery methods, including neural Granger causality~\cite{tank2021neural} and attention-based causal inference~\cite{assaad2022survey}, provide interpretable directional dependencies among time series. Although these approaches enhance explanatory power, they often assume static causal structures or require retraining when the underlying dynamics shift. Our approach complements causal modeling by enabling context-aware adaptation, allowing dependency structures to evolve in response to new observations, consistent with in-context learning paradigms~\cite{brown2020language, zhang2024largelanguagemodelstime}.

\textbf{NLP-inspired models.} Inspired by advances in NLP, recent studies adapt attention mechanisms and large language models to time series analysis, graph learning~\cite{wu2024can}, forecasting~\cite{jin2024position, carson2024llmabbaunderstandingtimeseries, xue2023promptcast}, and causal benchmarking~\cite{zhou2024causalbench}. Models such as UniTime~\cite{liu2024unitime} leverage language instructions for cross-domain generalization, while GPT4TS exploits frozen language models for multiple time series tasks~\cite{zhou2023one, wang2024chattime}. Despite their flexibility, these approaches rely on implicit representations that limit structural interpretability. In contrast, our method retains explicit temporal and relational structures while supporting adaptive, in-context forecasting.

\section{Detailed steps on CAARL} \label{caarl}

\subsection{Step-1: In-context scenario creation}
\subsubsection{Model identification} \label{models}
The procedure for identifying key models from which variations in the series can be approximated is outlined as follows:

\noindent \textbf{(1) Models initialization: } Considering $\mathcal{CS} = \{S^i,\, i\,=\,1,\,2,\, \ldots,\, n\}$ as an ensemble of co-evolving time series, at a specified initial time interval $T_1$, auto-regressive processes $f(T_1\,|\,\widehat{\Theta}^i,\,S^i)$ (as shown in Eq.~(\ref{Eq:autoregression})), $1\leq i\leq n$ are utilized to generate each sub-time-series $S^i_1$. By using a clustering function $Clust()$ we group the set of models $\widehat{\Theta} = \{\widehat{\Theta}^i, \, i\,=1,\, \ldots,\, n\}$ into distinct subsets of homogeneous models (i.e. models enabling the generation of the same sub-time-series) $\Theta = \{\Theta^{\kappa},\, 1\leq \kappa \leq K\}$ as follows,
\begin{align}\label{Eq:clustering-function}
    Clust(\Gamma\,|\, \widehat{\Theta}) &= \Theta\, ,
\end{align}
\noindent where $\Gamma$ is the set of parameters of the clustering function that can be approximated using the loss function given as,
\begin{align}\label{Eq:loss-function}
    \mathcal{L}(\Gamma) &= \sum_{\kappa=1}^K \sum_{l \in C_{\kappa}}  MSE\left(S^l_1,\, f(T_1\,|\,\Theta^{\kappa}, S^l)\right) + \notag \\
     &\;\;\;\;\;\;\;\;\;\;\;\;\;\;\;\;\;\;\;\;\;\;\;\;\;\;\;\;\;\;\;\;\;\;\;\;\;\;\;\;\;\;\;\;\;\;\;\;\;\; \tau \sum_{\kappa=1}^K Var(C_{\kappa}) \, ,
\end{align}
\noindent where $MSE()$ is the mean-square-error,  $C_{\kappa}$ represents the set of indices of models in the $\kappa$-th cluster. $\Theta^\kappa$ is the representative model the $\kappa$-th cluster, which minimizes the loss within its cluster. $\text{Var}(C_{\kappa})$ is the variance of the predictions within the $\kappa$-th cluster, calculated as the variance of the $MSE$ values for all models in $C_{\kappa}$ with respect to their generated sub-time-series. $\tau$ is a regularization parameter that controls the trade-off between the model fit and the homogeneity within each cluster (lower variance within clusters is preferred).

\noindent \textbf{(2) Models tracking:} Moving through time intervals, the process (\textbf{(1) Models initialization}) is repeated at each time interval $T_j$ ahead if and only if some sub-time-series exist for which none of the identified models can regenerate their values at that time interval. Formally, if at a current time interval $T_j > T_1$, $\exists \widehat{\Theta}^* \neq \widehat{\Theta}$ enabling the generation of some series $\overline{\mathcal{CS}} \subseteq \mathcal{CS}$ at this time interval $T_j$. We perform the same clustering strategy over parameters $\widehat{\Theta}^*$ as illustrated in \textbf{(1) Models initialization} and we update the final set $\Theta$ of models.

At the end of this pattern identification process, we will obtain a final set of patterns $\Theta = \{\Theta^{\kappa},\, 1\leq \kappa \leq K\}$ from which we can regenerate the whole set of time series $\mathcal{CS}$. In other words, given a set of co-evolving time series $\mathcal{CS}$ observed at different consecutive time intervals $T_1,\, T_2,\, \ldots,\, T_m$. At each time interval $T_j$, each sub-time-series $S^i_j$ variation of series $S^i \in \mathcal{CS}$ can be regenerated via a specific pattern $\Theta^{\kappa} \in \Theta$. Hence, for the whole set of time intervals $T_1,\, T_2,\, \ldots,\, T_m$, each time series $S^i$ can approximately be rewritten as a concatenation of autoregressive processes given as
\begin{align}\label{Eq:sequence-rewrote}
    S^i &= \odot_{j=1}^m S^i_j 
    \approx \odot_{j=1}^m \sum_{\kappa=1}^K \delta^{\kappa}_j \cdot f(T_j|\Theta^{\kappa}, S^i)\, ,
\end{align}
where $\delta^{\kappa}_j \in \{0,\, 1\}$ is a switch that takes value $1$ when the time series can be regenerated via the pattern $\Theta^{\kappa}$ at time interval $T_j$.

% Given the relationship outlined in Eq.~(\ref{Eq:sequence-rewrote}), the following lemma can be formulated:
% \begin{lemma}\label{lemma:dependency}
% Given two time series $S^1$ and $S^2$. If they can be successfully approximated by the same model $\Theta^{\kappa}$ at two consecutive time intervals, then there exists a dependency between these two time series at these intervals.
% \end{lemma}
% Lemma~\ref{lemma:dependency} suggests that if a single model successfully approximates two distinct time series across two consecutive time intervals, it demonstrates that the behavior of one series can offer insights into, or predict, the behavior of the other during those intervals. Building on this assertion from Lemma~\ref{lemma:dependency} the following theorem can be given, 

% \begin{theorem}\label{theorem:dependency}
% Consider the set $\mathcal{CS} = \{S^i, i = 1, \ldots, n\}$ representing $n$ time series observed across $m$ time intervals from $T_1$ to $T_m$. If there exists a finite set of models $\Theta = \{\Theta^{\kappa},\,1\leq \kappa \leq K\}$ that can accurately reconstruct these series over their respective intervals, and if $K < n \times m$, then it indicates that the time series in $\mathcal{CS}$ are dependent and thus are considered to co-evolve throughout the time intervals $T_1$ to $T_m$.
% \end{theorem}

\subsubsection{Series dependencies}
Based on the identified models $\Theta = \{\Theta^{\kappa}\,|\,\kappa=1,\ldots,K\}$, CAARL proceeds in building 
% we build 
a temporal graph $\mathcal{SG} = \{\mathbf{G}_{1,\, 2},\, \mathbf{G}_{2,\, 3}, \ldots, \mathbf{G}_{m-1,\,m} \}$ where each $\mathbf{G}_{j,\, (j+1)} = (\mathcal{N}_{\Theta}, E_{j,\, (j+1)},\, A_j, A_{j,j+1})$ is a bipartite, attributed and directed graph with $\mathcal{N}_{\Theta} = \{1, \ldots, K\}$ the set of nodes tagged by numerals representing each parameter $\Theta^{\kappa}$, $1\leq \kappa \leq K$ respectively. $E_{j,\, (j+1)}$ is the set of edges
%of oriented edges 
relating to the transition change that a time series may undergo in consecutive time intervals. It is given as,
\begin{align}\label{Eq:edge-relation}
    E_{j,\, (j+1)} &= \Bigl\{\left(\mu, \nu\right), \, \mu, \nu \in \mathcal{N}_{\Theta}\,|\, \exists S^i \in \mathcal{CS}, \Bigr. \notag \\
    &\Bigl. f(T_j|\Theta^{\mu})\approx S^i_j \,\land \, f(T_{j+1}|\Theta^{\nu})\approx S^i_{j+1}\Bigr\}\, .
\end{align}
and
\begin{align}\label{Eq:node-attributes}
% A_j &= 
% \Bigl \{v^{\mu}_j \in \{0,\, 1\}^n\,|\, \forall \nu,\, (\mu, \nu) \in E_{j, (j+1)}  \Bigr \}\, ,  A_{j+1} = \Bigl \{v^{\nu}_{j+1} \in \{0,\, 1\}^n\,|\,  \forall \mu,\, (\mu, \nu) \in E_{j, (j+1)} \Bigr \}\, ,
    \begin{cases}
        A_j = \Bigl \{v^{\mu}_j \in \{0,\, 1\}^n\,|\, \forall \nu,\, (\mu, \nu) \in E_{j, (j+1)}  \Bigr \}\, , \\
        A_{j+1} = \Bigl \{v^{\nu}_{j+1} \in \{0,\, 1\}^n\,|\,  \forall \mu,\, (\mu, \nu) \in E_{j, (j+1)} \Bigr \}\, ,
    \end{cases}
\end{align}
are the sets of outgoing and incoming node attributes. Each attribute dimension $v^{\mu, i}_j$ takes the value $1$ if the time series $S^i$ can be approximated using $\Theta^{\mu}$ at time interval $T_j$ and $0$ otherwise. 

\subsection{Reasoning over the temporal graph}

CAARL exploits the temporal graph $\mathcal{SG}$, which captures contextual dependencies among series $S^i \in \mathcal{CS}$ across intervals, to reason about their co-evolution. To make these dynamics interpretable, $\mathcal{SG}$ is serialized into natural language through a function $Serialized(S^i \,|\, q,\, \mathcal{SG})$. This function examines the current graph state $\mathbf{G}_{(j-1),j}$ and the $q-1$ preceding states to generate a narrative of transitions from $T_{j-q}$ to $T_{j-1}$, ultimately specifying the model expected to approximate $S^i$ at $T_j$. The serialization also clarifies whether a series switches between models across intervals.  

This graph-to-text process yields an explicit reasoning trace, transforming structural dependencies into narratives usable by Large Language Models (LLMs). Unlike prior serialization work focused on tabular data \cite{jaitly2023towards, hegselmann2023tabllm}, CAARL extends the idea to temporal graphs. In zero-shot settings, the serialized narrative guides the LLM to infer the next model; in few-shot mode, the LLM is fine-tuned to map narratives to the correct models with higher precision.

\subsection{Step-3: Prediction task}
The prediction task seeks to anticipate the changes in each series $S^i \in \mathcal{CS}$ at future, unseen time intervals, commonly referred to as time series forecasting. With a given set of time series $\mathcal{CS}$, their associated contextual scenario $\mathcal{SG}$, a defined list of models $\Theta = \{\Theta^{\kappa}\,|\,1\,\leq \,\kappa \,\leq\, K\}$, and the required last $q$ states of the temporal graph ($\mathbf{G}{(m-q),(m-q+1)}, \, \ldots,\,\mathbf{G}{(m-1),m} $.), the forecasting process unfolds through a hierarchical procedure outlined in the following algorithmic phases:

\noindent \textbf{(1) Model switch prediction: } During this phase, the prediction task focuses on determining whether a time series $S^i$, given the $q$ most recent states of the contextual scenario $\mathcal{SG}$, can still be approximated at the subsequent time interval $T_{m+1}$ by the same model or another model could be exploited,
will continue  exhibiting its current pattern at the subsequent time interval $T_{m+1}$ or switch to a different one.  Here, utilizing the graph serialization function, the information is passed to the LLM. The LLM's primary function is to predict whether the target time series can be approximated with another model or the current one. The output of this phase is a ``\textit{yes}'' or ``\textit{no}''.

\noindent \textbf{(2) Model prediction: } If the outcome of the initial phase indicates a ``\textit{no}'' for a given time series $S^i$, then the predicted model will be the same as the current model enabling the approximation of this series at the last time interval $T_m$. Conversely, if the response is ``\textit{yes}'', the LLM will proceed to generate the corresponding model $\widehat{\Theta}^{\kappa}$.

\noindent \textbf{(3) Series value forecasting: } Using the predicted model from phase \textbf{(2) Model prediction}, CAARL employs the current values of the target time series $S^i_m$ to forecast the subsequent values using the linear autoregressive process $f(T_{m+1},|,\widehat{\Theta}^{\kappa},, S^i)$.

\begin{table}[!t]
\centering
\caption{Data description. The dimension $|CS|$ indicates the number of time series, and $|\Theta|$ corresponds to the number of models achieved.}
\label{tab:data_description}
\resizebox{.5\textwidth}{!}{%
\begin{tabular}{|c|c|c|c|c|c|c|c|}
\hline
Dataset & $|CS|$ & \# Timestamps & Frequency & $|T_1|$ & Stride & $m$ & $|\Theta|$ from $T_1$ to $T_m$ \\
\hline
Cross rate & 25 & 4,175 & Daily & 194 & 97 & 42 & 19 \\
NYCT & 200 & 2,160 & Hourly & 201 & 100 & 20 & 64 \\
Rock & 70 & 2,844 & Hourly & 132 & 66 & 42 & 170 \\
S\&P 500 & 74 & 1,259 & Daily & 50 & 25 & 37 & 29 \\
\hline
\end{tabular}%
}
\end{table}

\section{Experiment} \label{experiments}
\subsection{Dataset Description and Experiment Setting}
In this section, we evaluate CAARL on four datasets spanning different domains: (i) \textit{Cross rate}, a collection of 25 foreign exchange rates; (ii) \textit{NYCT}, consisting of 200 New York City taxi trip time series\footnote{\url{https://www.data.cityofnewyork.us/Transportation/2018-Yellow-Taxi-Trip-Data/t29m-gskq}}; (iii) \textit{Rock}, comprising 70 geophysical rock series\footnote{\url{https://www.cs.ucr.edu/~eamonn/time_series_data_2018/}}; and (iv) \textit{S\&P 500}, which includes 74 large-cap IT sector stocks. Table~\ref{tab:data_description} summarizes the characteristics of these datasets, including the number of series ($|CS|$), sampling frequency, number of timestamps, initial interval size ($|T_1|$), stride, number of intervals ($m$), and the number of autoregressive models ($|\Theta|$) retained from $T_1$ to $T_m$. For evaluation, the last 45 time points of each series are withheld as a blind test set to assess forecasting performance.

\begin{figure*}[!t]
    \centering
    \subfloat[Two time series to approximate.]{\includegraphics[width=.9\textwidth]{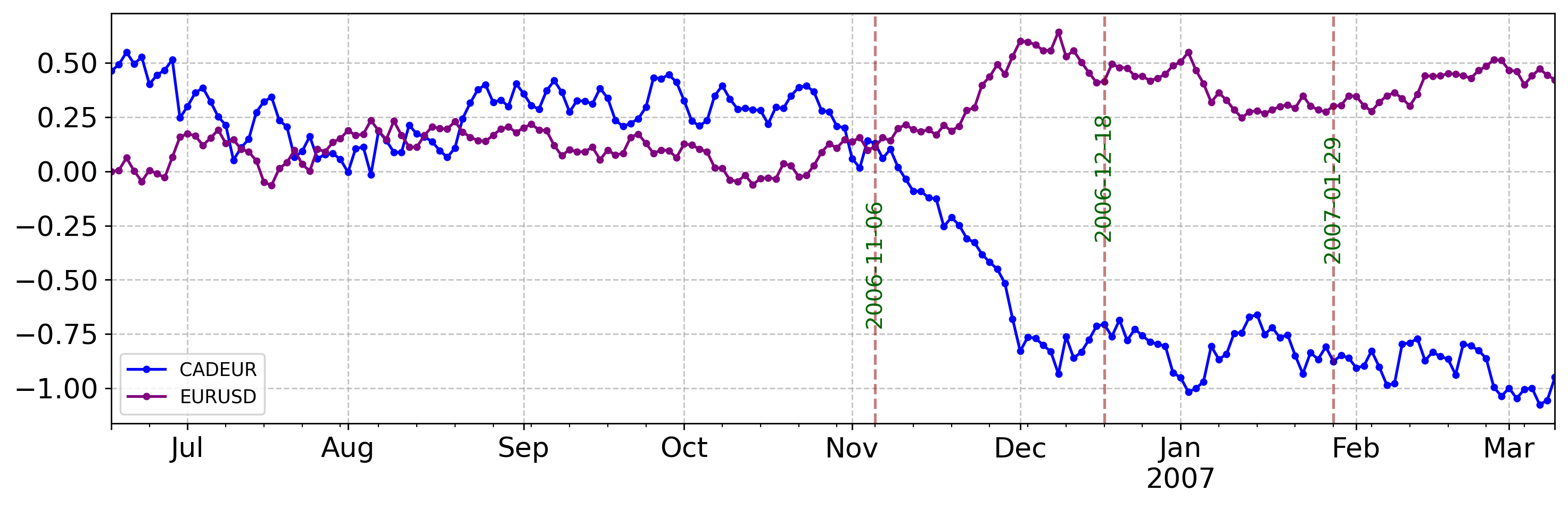}}\\
        \subfloat[2006-11-06 to 2006-12-18]{\includegraphics[width=.32\textwidth]{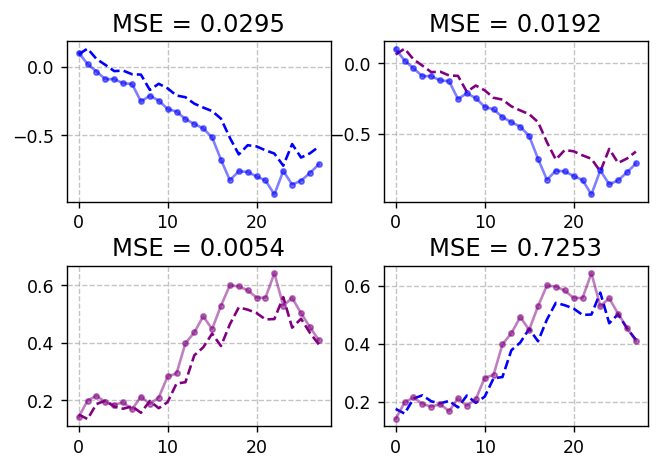}}
        \subfloat[2006-12-18 to 2007-01-29]{\includegraphics[width=.32\textwidth]{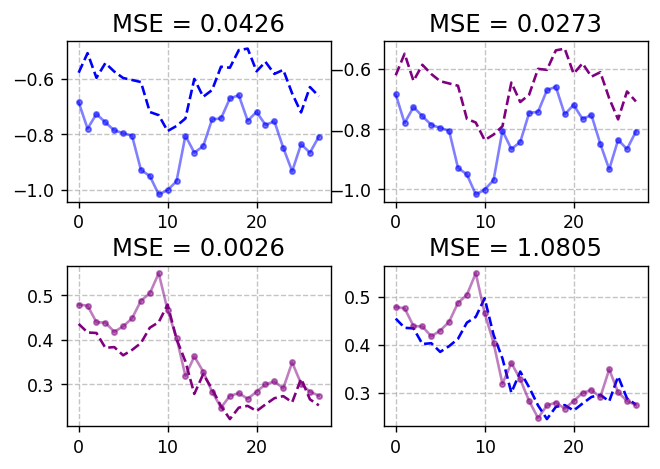}}
        \subfloat[2007-01-29 to 2007-03-05]{\includegraphics[width=.32\textwidth]{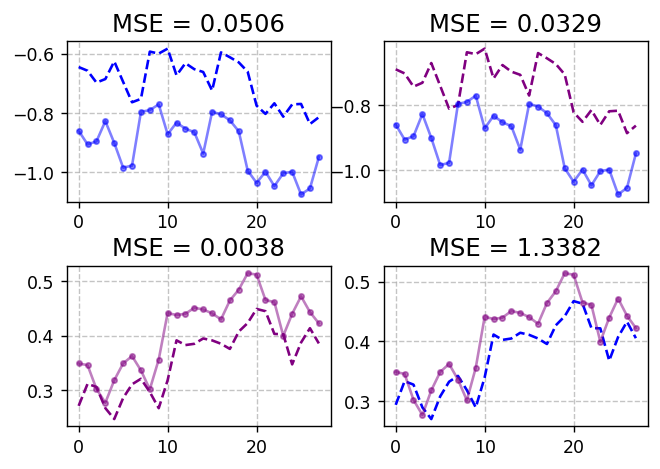}}
    \caption{Use case dependency illustration from Cross rate dataset.}
    \label{fig:fx-2-series-dependencies-representation}
\end{figure*}

In our analytical process, an autoregressive approach with a lag of \( p=2, 2, 4 \) is applied across all datasets to identify evolving patterns within the time series. Advanced language models, including OpenAI's GPT-3.5-Turbo \cite{ye2023comprehensive}, GPT-4 \cite{achiam2023gpt}, and DaVinci \cite{heryanto2023evaluating}, are also integrated into our models for linguistic analysis. The effectiveness of our framework is measured through the use of forecasting accuracy using Mean Absolute Error (MAE), Mean Squared Error (MSE) and Mean Absolute Percentage Error. Various baselines, including multivariate regressive models which utilize historical values from different time series as inputs, graph-based models such as RGNN \cite{seo2018structured}, DCRNN \cite{li2017diffusion}, and TGCN \cite{zhao2019t}, and transformer-based models such as Temporal Fusion Transformers (TFT) \cite{lim2020temporalfusiontransformersinterpretable}, DLinear \cite{zeng2023transformers} and Patch-TST (P-TST) \cite{nie2022time} are used to benchmark our models, demonstrating the robustness and precision of our approach in both few-shot and zero-shot scenarios.

\subsubsection{Model Identification}
Following the procedure described in Section~3.1, the first step of CAARL identifies autoregressive models that approximate the series values over successive intervals. As reported in Table~\ref{tab:data_description}, the initial interval lengths are $|T_1| = 194,\,201,\,132,$ and $50$ for the Cross rate, NYCT, Rock, and S\&P~500 datasets, respectively. The stride and interval sizes are automatically determined through a selection process that ensures stationarity within each segment (details provided in the Appendix).  

Across all four datasets, the number of distinct models $|\Theta|$ identified grows from the first interval $T_1$ to the final interval $T_m$, reflecting the inherent non-stationarity of the series. A further observation is that for every dataset, the product $|\mathcal{CS}| \times m$ is consistently larger than the number of discovered models $|\Theta|$. This indicates that multiple time series reuse the same autoregressive models across different intervals, thereby revealing latent dependencies and recurring regimes. Such reuse highlights the contextual interconnectedness of the series and emphasizes the dynamic but structured nature of their co-evolution.

\begin{figure*}[!t]
    \centering
\includegraphics[width=.9\textwidth]{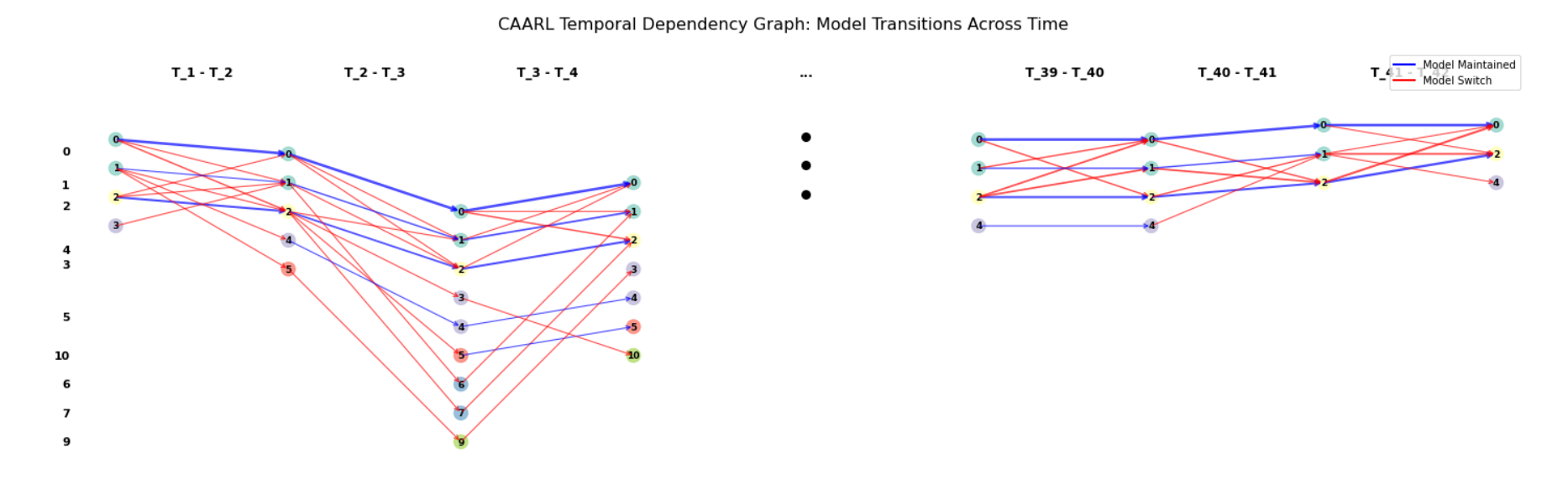}
    \caption{Example of dependencies in cross rate dataset.}
    \label{fig:in-context graph}
\end{figure*}

\subsubsection{Series dependencies}
By monitoring the likelihood of different models successfully approximating a series across successive time intervals, we can uncover insights into the dynamic switching behavior exhibited by the series. This analysis allows us to visualize and comprehend the transitions between models used to approximate the series, offering a deeper understanding of their evolving interactions. For example, consider Fig.~\ref{fig:fx-2-series-dependencies-representation}(a), which illustrates two time series, CADEUR (in blue) and EURUSD (in purple), from the Cross rate dataset. These series are analyzed using two models that respectively generate their values. When we evaluate these models beyond the date 2006-11-06, as shown in Fig.s~\ref{fig:fx-2-series-dependencies-representation}(b)-(d), it becomes evident that the model for CADEUR does not consistently generate its values accurately in subsequent intervals. For instance, in Fig.~\ref{fig:fx-2-series-dependencies-representation}(b), based on mean-square-error analysis, the plots indicate that the model for CADEUR (top left) is less accurate compared to the model for EURUSD (top right). Conversely, the model for EURUSD (bottom left) accurately reproduces its values, unlike the model for CADEUR (bottom right). This trend is even more apparent in later intervals shown in Fig.s~\ref{fig:fx-2-series-dependencies-representation}(c) and (d), where the model for CADEUR underperforms relative to the one for EURUSD. This overlapping data helps CAARL to establish a relationship between the two series, illustrating their interconnected dynamics.

In the given scenario, CAARL constructs a temporal dependency graph to visualize the evolution of model selection across sequential time intervals; this dynamic bipartite graph, illustrated in Fig.~\ref{fig:in-context graph}, maps transitions from $T_1$-$T_2$ through $T_{41}$-$T_{42}$ for the Cross Rate dataset, revealing a shifting interplay between models. The fluctuations in edge counts highlight the system's active switching between modeling approaches to accurately approximate changing series dynamics, while intervals of stability, evidenced by horizontal arrows, such as from $T_4$ to $T_5$ indicate where a single model was successfully maintained without adjustment, thus contrasting adaptive switching with periods of consistent capture.

\subsection{Graph serialization}
\begin{figure}[!t]
    \centering
    \includegraphics[width=.5\textwidth]{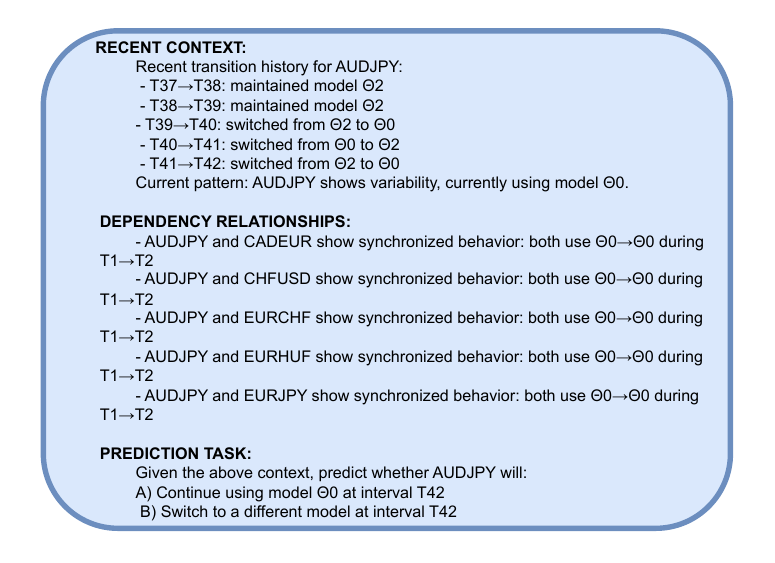}
    \caption{Serialization of graphs. Examples of series AUDJPY tracked within the cross rate temporal graph structure.}
    \label{fig:example of serialization}
\end{figure}

For serialization purposes, we develop textual descriptions that trace the progression of each series based on the models that approximate their values over time. In Fig.~\ref{fig:example of serialization}, we showcase an example of such trajectory, drawing from a time series AUDJPY chosen from the Cross rate dataset. By examining the variation over consecutive time intervals, we assess the ability of different models to approximate the series. In this case for instance,
where each time interval represents 194 days, the AUDJPY series can be modeled using two different models $\Theta^2$ and $\Theta^0$. Considering the dependency with other time series, the LLM will attempt predicting the subsequent suitable model(s) to use for forecasting purpose.  

\subsection{Forecasting}
To assess the efficacy of the approach, we explored its capacity to predict series values under two primary scenarios. In the first scenario, we utilized a serialization that spans five consecutive graph states (graph lag $q=5$), where the LLM predicts the next model capable of generating future series values without any prior training. This is referred to as zero-shot learning because the LLM is not pre-trained with specific labels. In the second scenario, known as few-shot learning, only a subset of data is used to train the LLM to recognize and map corresponding labels. We compared CAARL's performance against other models in these scenarios, as detailed in Table~\ref{tab:forecast_1}. According to the results shown in Table~\ref{tab:forecast_1}, CAARL typically forecasts data with high accuracy, comparable to transformer-based methods. Interestingly, CAARL performs better in the zero-shot scenario than in few-shot learning, highlighting its robustness in situations with limited training data.
\begin{table*}[!th]
    \centering
    \caption{Comparative forecasting. The best values are in green colour and the second best in orange colour. In the zero-shot, the LLM selects models for forecasting without any additional fine-tuning, relying solely on its pre-trained knowledge to interpret serialized time series trajectories and determine the most suitable model. In contrast, in the few-shot, the LLM undergoes fine-tuning using a small portion of the dataset (trajectories), enabling it to learn from the provided patterns and improve its ability to forecast or select models based on observed dependencies. }
    \label{tab:forecast_1}
\resizebox{1\textwidth}{!}{
    \begin{tabular}{|c|c|c|c|c|c|c|c|c|c|c|c|c|c|}
     \hline
    \multicolumn{14}{c}{\bf One time interval ahead} \\ \hline
        &&\multicolumn{3}{|c|}{Cross rate} & \multicolumn{3}{c|}{NYCT} & \multicolumn{3}{c|}{Rock} & \multicolumn{3}{c}{S\&P 500} \\ \hline
        \hline
       \multicolumn{14}{|c|}{Zero-shot} \\ 
       \hline
       \hline
       & Model & MAE & MSE & MAPE & MAE & MSE & MAPE & MAE & MSE & MAPE & MAE & MSE & MAPE \\
       \hline
       \hline
       \multirow{3}{*}{CAARL}
       & GPT3  & $\cellcolor{orange!25}\bf{0.36}$ & $0.24$ & $\cellcolor{green!25}\bf 0.34$ & $\cellcolor{green!25}\bf 0.72$ & $1.10$ & $8.44$ & $\cellcolor{green!25}\bf 0.15$ & \cellcolor{green!25}{$\bf 0.06$} & $\cellcolor{green!25}\bf 0.98$ & $0.38$ & $0.28$ & \cellcolor{green!25}{$\bf 0.89$}\\ 
       & GPT4 & $\cellcolor{orange!25}\bf{0.36}$ & $0.24$ & $\cellcolor{green!25} \bf0.34$ & $\cellcolor{green!25}\bf 0.72$ & $1.08$ & $9.25$ & $\cellcolor{orange!25}\bf 0.16$ & $\cellcolor{green!25}\bf 0.06$ & $1.05$ & \cellcolor{orange!25}{$\bf 0.37$} & $0.28$ & \cellcolor{orange!25}{$\bf 0.94$}\\
        & DAVINCI & $0.37$ & $0.27$ & $\cellcolor{orange!25}\bf 0.36$ & $\cellcolor{green!25}\bf 0.72$ & $1.08$ & $8.46$ & $\cellcolor{orange!25}\bf 0.16$ & $\cellcolor{green!25}\bf 0.06$ & $1.07$ & $0.38$ & $0.28$ & $0.95$\\ 
       \hline
        \hline
       \multicolumn{14}{|c|}{Few-shot} \\ \hline
       \hline
       \multirow{3}{*}{CAARL} 
       & GPT3  & $\cellcolor{orange!25}\bf 0.36$ & $\cellcolor{orange!25}\bf 0.23$ & $1.49$ & $\cellcolor{orange!25}\bf 0.74$ & $0.99$ & $24.94$ & $0.21$ & $\cellcolor{orange!25}\bf 0.11$ & $3.41$ & $0.38$ & $0.24$ & $1.37$ \\ 
       
       & GPT4  & $0.38$ & $0.24$ & $1.56$ & $\cellcolor{orange!25}\bf 0.74$ & $0.99$ & $25.54$ & $0.21$ & $\cellcolor{orange!25}\bf 0.11$ & $3.41$ & $0.38$ & \cellcolor{orange!25}{$\bf 0.23$} & $1.38$ \\ 
       & DAVINCI  & $0.38$ & $\cellcolor{orange!25}\bf 0.23$ & $1.49$ & $\cellcolor{orange!25}\bf 0.74$ & $0.99$ & $26.01$ & $0.21$ & $\cellcolor{orange!25}\bf 0.11$ & $3.41$ & $0.38$ & \cellcolor{orange!25}{$\bf 0.23$} & $1.31$\\ 
       \hline
       \multirow{3}{*}{\makecell{Multivariate}}
       & VAR  & $++$ & $++$ & $++$ & $++$  & $++$ & $\cellcolor{orange!25}\bf 2.26$ & $++$ & $++$ & $3.21$ & $++$ & $++$ & $2.91$\\  
       & SVR   & $1.18$ & $2.37$ & $11.0$ & $++$ & $++$ & $\cellcolor{orange!25}\bf 2.26$ & $4.30$ & $29.0$ & $3.22$ & $6.40$ & $57.6$ & $2.89$ \\ 
       
       & GPR   & $1.18$ & $2.37$ & $11.0$ & $++$ & $++$ & $\cellcolor{orange!25}\bf 2.26$ & $4.30$ & $29.0$ & $3.22$ & $6.40$ & $57.6$ & $2.89$ \\ 
       \hline
       \multirow{3}{*}{Graph-based}
       % & RGCNN & $0.90$ & $1.09$ & $++$ & $0.93$ & $1.54$ & $++$ & $0.22$ & $\cellcolor{orange!25}\bf 0.11$ & $++$ & $0.88$ & $1.10$ & $++$\\ \cline{2-14}
       & DCRNN  & $0.98$ & $1.28$ & $++$ & $0.88$ & $1.41$ & $++$ & $0.30$ & $0.17$ & $++$ & $0.77$ & $0.92$ & $++$\\ 
       
       & TGCNN  & $1.06$ & $1.56$ & $++$ & $0.91$ & $1.50$ & $++$ & $0.25$ & $0.13$ & $++$ & $0.83$ & $1.01$ & $++$\\ 
       \hline
       \multirow{7}{*}{Transformers}
       & TFT  & $\cellcolor{green!25}\bf 0.20$ & $\cellcolor{green!25}\bf 0.08$ & $1.24$ & $0.85$ & $1.32$ & $4.22$ & $0.45$ & $0.31$ & $0.85$ & $\cellcolor{green!25}\bf 0.30$ & $\cellcolor{green!25}\bf 0.21$ & $4.58$\\
       
       & DLinear  & $0.65$ & $0.56$ & $1.00$ & $\cellcolor{orange!25} \bf 0.74$ & $\cellcolor{green!25} \bf 0.92$ & $\cellcolor{green!25} \bf 1.00$ & $0.75$ & $\bf 0.98$ & \cellcolor{orange!25}{$\bf 1.00$} & $0.65$ & $0.50$ & $1.00$\\ 
       
       & P-TST & $0.64$ & $0.54$ & $1.00$ & $0.75$ & $\cellcolor{orange!25} \bf 0.98$ & $\cellcolor{green!25} \bf 1.00$ & $0.73$ & $0.69$ & $1.01$ & $0.67$ & $0.53$ & $1.00$\\ 
       & Informer  & $0.99$ & $0.88$ & $1.46$ & $ 1.12$ & $ 1.38$ & $1.54$ & $1.15$ & $\bf 1.52$ & $1.55$ & $0.98$ & $0.76$ & $1.48$\\
       & Autoformer  & $0.88$ & $0.76$ & $1.34$ & $ 1.02$ & $ 1.25$ & $ 1.37$ & $1.02$ & $\bf 1.31$ & $ 1.37$ & $0.88$ & $0.69$ & $1.34$\\
       & FEDformer  & $0.82$ & $0.71$ & $1.27$ & $ 0.94$ & $ 1.15$ & $1.29$ & $0.97$ & $ 1.24$ & $ 1.29$ & $0.81$ & $0.64$ & $1.27$\\
       & TimesNet  & $0.75$ & $0.64$ & $1.17$ & $0.89$ & $ 1.16$ & $1.19$ & $0.87$ & $0.82$ & $1.19$ & $0.80$ & $0.63$ & $1.18$\\
       \hline
    \hline
    \multicolumn{14}{c}{\bf Four time intervals ahead} \\ \hline
       % &  & \multicolumn{3}{c|}{Cross rate} & \multicolumn{3}{c|}{NYCT} & \multicolumn{3}{c|}{Rock} & \multicolumn{3}{c}{S\&P 500} \\ \hline
       %  \hline
       \multicolumn{14}{|c|}{Zero-shot} \\ 
       \hline
       \hline
       & Model & MAE & MSE & MAPE & MAE & MSE & MAPE & MAE & MSE & MAPE & MAE & MSE & MAPE \\
       \hline
       \hline
       \multirow{3}{*}{CAARL}  & GPT3  & $0.34$ & $\cellcolor{orange!25} \bf 0.22$ & $\cellcolor{green!25} \bf 0.31$ & $\cellcolor{green!25} \bf 0.73$ & $1.10$ & $7.49$ & $0.15$ & $0.05$ & $\cellcolor{orange!25} \bf 0.79$ & $0.61$ & $0.70$ & $1.60$\\ 
       & GPT4  & $\cellcolor{orange!25} \bf 0.33$ & $\cellcolor{orange!25} \bf 0.22$ & $\cellcolor{green!25} \bf 0.31$ & $\cellcolor{green!25} \bf 0.73$ & $1.10$ & $6.72$ & $0.15$ & $0.05$ & \cellcolor{green!25}{$\bf 0.77$} & $0.61$ & $0.70$ & $1.60$\\
        & DAVINCI & $0.34$ & $\cellcolor{orange!25} \bf 0.22$ & $\cellcolor{green!25} \bf 0.31$ & $\cellcolor{green!25} \bf 0.73$ & $1.10$ & $6.61$ & $0.15$ & $0.05$ & $0.80$ & $0.61$ & $0.70$ & $1.60$\\ 
       \hline
        \hline
       \multicolumn{14}{|c|}{Few-shot} \\ \hline
       \hline
       \multirow{3}{*}{CAARL}  & GPT3  & $\cellcolor{orange!25} \bf 0.33$ & $\cellcolor{green!25} \bf 0.21$ & $\cellcolor{orange!25} \bf 0.33$ & $0.76$ & $\cellcolor{green!25} \bf 0.93$ & $8.28$ & $0.14$ & \cellcolor{orange!25}{$\bf 0.04$} & $1.23$ & \cellcolor{green!25}{$\bf 0.37$} & \cellcolor{green!25}{$\bf 0.24$} & \cellcolor{orange!25}{$\bf 0.93$} \\ 
       & GPT4  & $\cellcolor{green!25} \bf 0.32$ & $\cellcolor{green!25} \bf 0.21$ & $\cellcolor{orange!25} \bf 0.33$ & $0.76$ & $\cellcolor{orange!25} \bf 0.94$ & $8.28$ & \cellcolor{green!25}{$\bf 0.12$} & \cellcolor{green!25}{$\bf 0.03$} & $0.98$ & $\cellcolor{green!25} \bf 0.37$ & \cellcolor{orange!25}{$\bf 0.25$} & \cellcolor{green!25}{$\bf 0.92$} \\ 
       & DAVINCI  & $\cellcolor{orange!25} \bf 0.33$ & $\cellcolor{orange!25} \bf 0.22$ & $0.34$ & $0.76$ & $\cellcolor{orange!25} \bf 0.94$ & $8.28$ & $\cellcolor{orange!25} \bf 0.13$ & $0.04$ & $1.21$ & $\cellcolor{green!25} \bf 0.37$ & $0.24$ & $0.93$\\ 
       \hline
       \multirow{3}{*}{Multivariate} & VAR  & $++$ & $++$ & $22.0$ & $++$ & $++$ & $2.30$ & $++$ & $++$ & $1.00$ & $++$ & $++$ & $2.55$\\  
       & SVR   & $1.25$ & $2.74$ & $22.0$ & $++$ & $++$ & $2.30$ & $++$ & $++$ & $1.00$ & $3.10$ & $13.1$ & $2.51$ \\ 
       & GPR   & $1.25$ & $2.74$ & $22.1$ & $++$ & $++$ & $2.30$ & $++$ & $++$ & $1.00$ & $3.10$ & $13.1$ & $2.51$ \\ 
       \hline
       \multirow{3}{*}{ Graph-based} & RGCNN & $1.04$ & $1.47$ & $++$ & $0.83$ & $1.18$ & $++$ & $0.23$ & $0.13$ & $++$ & $0.66$ & $0.77$ & $++$\\ 
       & DCRNN  & $1.04$ & $1.47$ & $++$ & $0.84$ & $1.22$ & $++$ & $0.20$ & $0.09$ & $++$ & $0.66$ & $0.78$ & $++$\\ 
       & TGCNN  & $1.01$ & $1.38$ & $++$ & $0.84$ & $1.22$ & $++$ & $0.21$ & $0.11$ & $++$ & $0.65$ & $0.77$ & $++$\\ 
       \hline
       \multirow{7}{*}{ Transformers} & TFT  & $0.68$ & $0.64$ & $1.06$ & $\cellcolor{orange!25} \bf 0.74$ & $0.96$ & $\cellcolor{orange!25} \bf 1.02$ & $0.60$ & $0.53$ & $2.70$ & $\cellcolor{orange!25} \bf 0.59$ & $0.53$ & $1.04$\\
       & DLinear  & $0.66$ & $0.62$ & $1.00$ & $\cellcolor{orange!25} \bf 0.74$ & $0.96$ & $\cellcolor{green!25} \bf 1.00$ & $0.60$ & $0.53$ & $1.00$ & $\cellcolor{orange!25} \bf 0.59$ & $0.53$ & $1.00$\\
       & P-TST & $0.69$ & $0.66$ & $0.99$ & $\cellcolor{orange!25} \bf 0.74$ & $0.96$ & $\cellcolor{green!25} \bf 1.00$ & $0.60$ & $0.53$ & $1.01$ & $\cellcolor{orange!25} \bf 0.59$ & $0.53$ & $1.00$\\
       & Informer  & $0.95$ & $0.88$ & $1.42$ & $ 1.07$ & $1.33$ & $ 1.42$ & $0.86$ & $0.75$ & $1.43$ & $ 0.84$ & $0.75$ & $1.42$\\
       & Autoformer  & $0.89$ & $0.85$ & $1.35$ & $ 1.02$ & $1.30$ & $ 1.37$ & $0.82$ & $0.71$ & $1.36$ & $ 0.81$ & $0.71$ & $1.37$\\
       & FEDformer  & $0.84$ & $0.79$ & $1.27$ & $0.96$ & $1.21$ & $ 1.28$ & $0.76$ & $0.68$ & $1.26$ & $0.75$ & $0.67$ & $1.27$\\
       & TimesNet  & $0.82$ & $0.79$ & $1.16$ & $ 0.89$ & $1.14$ & $ 1.19$ & $0.72$ & $0.64$ & $1.18$ & $ 0.71$ & $0.63$ & $1.17$\\
       \hline
    \end{tabular}}
\end{table*}

\section{Conclusion}
This paper presents our novel approach, CAARL, to unravel the underlying in-context dynamics that drive changes in co-evolving time series. We employ a three-step methodology that adapts large language models, beginning with the formulation of a temporal graph to model the influences exerted by one time series on another, followed by serializing the graph into natural language text, and finally, applying a few-shot learning process to forecast subsequent variations across different timestamps. Our approach also sheds light on the driving factors (i.e., specific series) behind these changes. The approach is flexible, allowing any NLP-based model to generate subsequent values in a time series trajectory by leveraging its pre-trained knowledge. By encoding time series data as serialized text, these models can infer patterns and predict future values without requiring traditional time series architectures. This suggests a generalized framework where NLP-driven sequence modelling can be effectively repurposed for time series forecasting, expanding the applicability of language models beyond their conventional domains. Through experiments on real-world co-evolving time series datasets, our approach demonstrates superior performance over existing time series methods in capturing dependencies within co-evolving series.

% \appendices
% \section{Proof of the First Zonklar Equation}
% Appendix one text goes here.

% % you can choose not to have a title for an appendix
% % if you want by leaving the argument blank
% \section{}
% Appendix two text goes here.

% use section* for acknowledgment
\section*{Acknowledgment}
This work was supported by the Natural Sciences and Engineering Research Council of Canada (NSERC). ChatGPT was used solely to assist with language editing and text shortening. The authors are fully responsible for the scientific content.

% Can use something like this to put references on a page
% by themselves when using endfloat and the captionsoff option.
\ifCLASSOPTIONcaptionsoff
  \newpage
\fi

% trigger a \newpage just before the given reference
% number - used to balance the columns on the last page
% adjust value as needed - may need to be readjusted if
% the document is modified later
%\IEEEtriggeratref{8}
% The "triggered" command can be changed if desired:
%\IEEEtriggercmd{\enlargethispage{-5in}}

% references section

% can use a bibliography generated by BibTeX as a .bbl file
% BibTeX documentation can be easily obtained at:
% http://mirror.ctan.org/biblio/bibtex/contrib/doc/
% The IEEEtran BibTeX style support page is at:
% http://www.michaelshell.org/tex/ieeetran/bibtex/
%\bibliographystyle{IEEEtran}
% argument is your BibTeX string definitions and bibliography database(s)
% \bibliography{IEEEabrv,../bib/paper}
%
% <OR> manually copy in the resultant .bbl file
% set second argument of \begin to the number of references
% (used to reserve space for the reference number labels box)

\bibliographystyle{ieeetr}
\bibliography{references}

\end{document}